\documentclass[journal=jacsat,manuscript=article]{achemso}

\usepackage[version=3]{mhchem} 
\usepackage{multirow}
\usepackage{makecell}
\usepackage{cellspace}
\usepackage{booktabs}
\usepackage{amsmath} 
\usepackage{amssymb} 
\usepackage{enumitem}
\usepackage{tabularx}

\author{Janghoon Ock}
\email{jock@andrew.cmu.edu}
\affiliation[1]{Department of Chemical Engineering, Carnegie Mellon University, 5000 Forbes Street, Pittsburgh, PA 15213, USA}
\alsoaffiliation[2]{Toyota Research Institute, 4440 El Camino Real, Los Altos, CA 94022, USA}
\author{Joseph Montoya}
\email{joseph.montoya@tri.global}
\author{Daniel Schweigert}
\email{daniel.schweigert@tri.global}
\author{Linda Hung}
\email{linda.hung@tri.global}
\author{Santosh K. Suram}
\email{santosh.suram@tri.global}
\author{Weike Ye}
\affiliation[2]{Toyota Research Institute, 4440 El Camino Real, Los Altos, CA 94022, USA}
\email{weike.ye@tri.global}

\title[An \textsf{achemso} demo]
  {UniMat: Unifying Materials Embeddings through Multi-modal Learning}

\abbreviations{IR,NMR,UV}

\begin{document}



\begin{abstract}
Materials science datasets are inherently heterogeneous and are available in different modalities such as characterization spectra, atomic structures, microscopic images, and text-based synthesis conditions. The advancements in multi-modal learning, particularly in vision and language models, have opened new avenues for integrating data in different forms. In this work, we evaluate common techniques in multi-modal learning (alignment and fusion) in unifying some of the most important modalities in materials science: atomic structure, X-ray diffraction patterns (XRD), and composition. We show that structure graph modality can be enhanced by aligning with XRD patterns. Additionally, we show that aligning and fusing more experimentally accessible data formats, such as XRD patterns and compositions, can create more robust joint embeddings than individual modalities across various tasks. This lays the groundwork for future studies aiming to exploit the full potential of multi-modal data in materials science, facilitating more informed decision-making in materials design and discovery.
\end{abstract}

\section{Introduction}
Materials science datasets are inherently heterogeneous and are typically represented using modalities such as characterization spectra, atomic structure, micro-structure images, and text-based processing information, etc \cite{Kim2017, Zhao2022, Jensen2023, moro_multimodal_2024}. Each modality possesses fundamentally varying sensitivities to different characteristics of a material. Integration of information across multiple modalities is necessary to achieve a comprehensive understanding of a material. However, this integration has been a challenge due to the heterogeneity of the data and embedded information \cite{moro_multimodal_2024, gong_multimodal_nodate}. These datasets, typically collected using diverse methods, often differ significantly in their format and volume, which complicates their thorough integration. Despite these challenges, effective integration of multi-modal data can enable tasks such as enhanced property prediction, and retrieval/generation of one modality from other prior-known modalities. These tasks are expected to play a pivotal role in identifying the best materials to investigate and to identify the best materials characterization workflow for accelerated materials discovery. Wherein, accelerated materials discovery is needed to address grand challenges in areas such as environment, energy, and security. 

The advancements in multi-modal AI methods, driven by the success of vision and language models, have opened new avenues for combining diverse data formats. Several multi-modal AI methodologies \cite{clip, ramesh2022, girdhar2023, xue2023} have emerged recently, and are currently being used to integrate various modalities in practical applications such as retrieval of one modality from another, for example, a text description into credible images or song lyrics into musical accompaniment. These methodologies also enable enhancement of representations learned for the single modalities, i.e. the integration of textual descriptions with visual data improves the accuracy and contextual appropriateness of image captions \cite{nguyen2023, ji2023}. 

In addition, single-modal machine learning approaches in materials science have significantly progressed. Methods that use chemical composition, crystal structure, and various spectroscopic techniques in isolation have matured significantly, making them viable building blocks for materials discovery and design campaigns. Thus, it is an opportune time for us to adapt advancements in multi-modality to the materials science domain.

The adaptation should be grounded in a few objectively verifiable metrics. Particularly, we judge the relevance of multi-modal AI methods in materials science based on the following criteria : 
\begin{itemize}
    \item Can we enhance the predictive capability of one modality through integration of additional modalities?
\end{itemize}
\begin{itemize}
    \item Can experimentally obtainable modalities be integrated to obtain the information available via simulated modalities? 
    This criterion has particular relevance for real world applications where-in information such as crystal structure typically available via simulation modalities are difficult to access through experiments preventing accelerated materials discovery
\end{itemize}
\begin{itemize}
    \item Can modalities with inherently weaker sensitivity to certain features benefit other modalities? 
\end{itemize}
While prior approaches have reported on sub-parts of these criteria, our work focuses on addressing all of these fundamental criteria using multi-modal AI methods. We particularly emphasize on the utility of multi-modal AI methodologies to capture material properties from experimentally accessible modalities.

In this contribution, we present UniMat, a multi-modal AI methodology that unifies material embeddings from various modalities, such as XRD patterns, structure graphs, and compositional features. We demonstrate that lattice lengths and angles can be better predicted with crystal structure graphs aligned with simulated XRD patterns, compared to crystal structure graphs alone. By fusing and aligning embeddings from the XRD patterns and chemical composition, we achieve predictive accuracy comparable to models that rely solely on structure graphs, outperforming single-modality approaches like XRD-only or composition-only models. Our comparison of XRD/composition-based predictions with structure graph-based methods also demonstrates how experimental modalities (XRD and composition) can be effectively combined to achieve performance on par with models that strictly rely on crystal structure data.

\section{Methods}
\subsection{Structure graph modality}

In this work, structure as a modality in materials science refers to the arrangement of atoms within a material (atomic coordinates). This has typically been represented as graphs where nodes represent atoms and edges represent bonds or interactions between them; Graph Neural Networks (GNNs) \cite{cgcnn, chmiela2017machine, gasteiger_fast_2022} are used to encode these structure graphs by learning representations that capture the connectivity and properties of the nodes and edges to predict various material properties. In our study, we chose DimeNet++ as the encoder.  

\subsection{XRD modality} 
  
X-ray diffraction (XRD) is a widely used characterization technique that provides a unique fingerprint of a material's atomic structure by analyzing the interference patterns emerging from X-rays scattered by the atoms within the material. The resulting XRD spectrum is typically represented as a 2D array of intensity (\textit{I}) versus diffraction angle ($2\theta$), where the peaks correspond to the specific crystal planes. 

In this study, we used simulated XRD patterns derived from atomic structures \cite{cao2024, pymatgen}, represented as stick patterns, and applied Gaussian smearing to align the intensities on an evenly spaced $2\theta$ grid. These processed intensities were then encoded as a 1D array, suitable for input into machine learning models. Specifically, we employed Convolutional Neural Networks (CNNs) to learn the characteristic peaks and patterns of the XRD using convolutional filters.

\subsection{Composition modality}

The composition of a material is a modality that represents the types of the constituent elements and their ratios. Methods to represent compositional information typically include one-hot encoding or vectors of numerical elemental properties such as atomic number, electronegativity, etc. In this work, we used the Magpie features \cite{ward2016general}, a typical example of the latter, to represent the composition, followed by a multilayer perceptron (MLP) to further encode the representation.

\subsection{Alignment}

Embeddings from different modalities should be aligned in the latent space to enable their effective integration. This alignment involves increasing the similarity between embeddings corresponding to the same material, while ensuring that embeddings from different materials are maximally dissimilar. Without such alignment, embeddings from different modalities may exist in non-overlapping spaces, hindering the ability to fully leverage the complementary information each modality offers. To achieve this, we implement a contrastive loss between two different modalities, inspired by CLIP \cite{clip}. 

The mathematical formulation of the contrastive loss function is defined as follows:

\begin{equation}
\mathcal{L} = -\frac{1}{N} \sum_{i=1}^{N} \log \frac{e^{\text{sim}(z_{m_1}^i, z_{m_2}^i)/\tau}}{\sum_{j=1}^{N} e^{\text{sim}(z_{m_1}^i, z_{m_2}^j)/\tau} \mathbb{I}_{\{i \neq j\}}} - \frac{1}{N} \sum_{i=1}^{N} \log \frac{e^{\text{sim}(z_{m_2}^i, z_{m_1}^i)/\tau}}{\sum_{j=1}^{N} e^{\text{sim}(z_{m_2}^i, z_{m_1}^j)/\tau} \mathbb{I}_{\{i \neq j\}}}
\end{equation}

In this expression, \( z_1^i \) and \( z_2^i \) represent the embeddings of the i-th entry of modality 1 and 2, respectively. Additionally, \( \tau \) is a temperature parameter that scales the similarity scores within the model. We use a batch size as \(N\) in our training architecture. The similarity function,  \( \text{sim} \), calculates the cosine similarity between the embeddings as follows:

\begin{equation}
\text{sim}(z_{m_1}, z_{m_2}) = \frac{z_{m_1} \cdot z_{m_2}}{\|z_{m_1}\| \|z_{m_2}\|}
\end{equation}

In this study, we apply this alignment to pairs of structure graphs and XRD patterns, and pairs of XRD patterns and compositions.

\subsection{Fusion}

To effectively utilize different modalities, fusing is another important approach. Among the various fusion strategies, the simplest is to concatenate embeddings. An active area of research is late-fusion methods, which combine embeddings from the latter layers due to their ability to integrate information from different modalities at a more advanced stage of processing \cite{tziafas2023, yu2024}. Here, we focus on establishing a baseline with the simpler concatenation approach. In this study, we concatenated the embeddings from either the last readout layers of the GNN and CNN or the last layer of the MLP, and followed by an MLP fusion block consisting of three linear layers. Refer to Figure \ref{fig:fusion} for a schematic illustration of the fusion blocks.

\subsection{Downstream Tasks}

We evaluate our embedding unification method across both regression and classification tasks. For regression, we predict the material properties of formation energy (eV/atom) and lattice parameters (length in Å and angles in degrees. The classification task involves categorizing the input into seven crystal systems present in materials.

Formation energy is an important property for inorganic crystals as it provides insight into the material's stability, representing the energy change that occurs when a compound forms from its constituent elements in their reference states \cite{Bartel2018, Bartel2022}. Lattice parameters and crystal system capture the symmetry of the lattice, which describes the specific ordered arrangement of atoms in the crystal structure that repeats periodically in three-dimensional space. Together, these tasks target fundamental properties and parameters of inorganic crystals.

These tasks are accomplished by adding MLP layers on top of the encoders. The embeddings passed to the regression or classification MLP layers can be single, aligned, and/or fused embeddings. 

\subsection{Data setup}

The dataset used in this study is the MP20 subset of the Materials Project (MP) database \cite{jain2013commentary}. MP20 consists of 45,231 inorganic crystal materials from the Inorganic Crystal Structure Database (ICSD), the largest database of fully characterized inorganic crystal structures, and have less than 20 atoms in the unit cell. We selected this dataset because it has been widely used as a benchmark for various models\cite{xie_crystal_2022, gasteiger_fast_2022}. We adopted the data split setup from Xie et al., which employs a 60-20-20 random split.

The MP20 dataset includes the atomic coordinates (x, y, z) for each crystal structure, as well as key properties such as formation energy. The lattice parameters, the composition, and the crystal systems can be obtained and/or determined directly from the crystal structure. The simulated XRD patterns are simulated using the Pymatgen library \cite{pymatgen}. Gaussian smearing (\(\sigma\) = 0.3\textdegree) is applied to align all the patterns to a consistent 2\(\theta\) grid (2\(\theta\) = 0-90\textdegree).

\subsection{Related work}
Recent advancements in multi-modal machine learning methodologies have shown significant improvements in predicting material properties and enhancing machine learning performance. COSNet\cite{gong_multimodal_nodate}, a composition-structure bimodal network, demonstrates reduced prediction errors for material properties by effectively utilizing incomplete structure information and data augmentation with an attention fusion block \cite{gong_multimodal_nodate}. Similar works, such as CrysMMNet \cite{das_crysmmnet_2023} and CLICS \cite{ozawa_graph-text_2024}, focus on graph-text pre-training. CrysMMNet demonstrates that concatenation of graph and language embeddings is superior compared to sum and average methods. CLICS shows that CLIP-style alignment between a pre-trained GNN and a language encoder trained on materials science literature increases classification accuracies for space groups and local atomic environments. Moro et al.\cite{moro_multimodal_2024} expanded the modalities by integrating natural language descriptions, graph representations, density of states (DOS), and charge density.

Inspired by these strategies and the guiding criteria laid out in the Introduction section, our work aims to distinguish itself by leveraging experimentally accessible modalities, such as characterization patterns themselves, to potentially circumvent the often encountered bottleneck of solving for structure. This approach enhances the practical applicability of our methodology in real-world scenarios.

\section{Experiment and Result}
\subsection{Alignment increases GNN-based prediction accuracy}

\begin{figure}[!hptb] 
  \centering
  \includegraphics[width=0.8\linewidth]{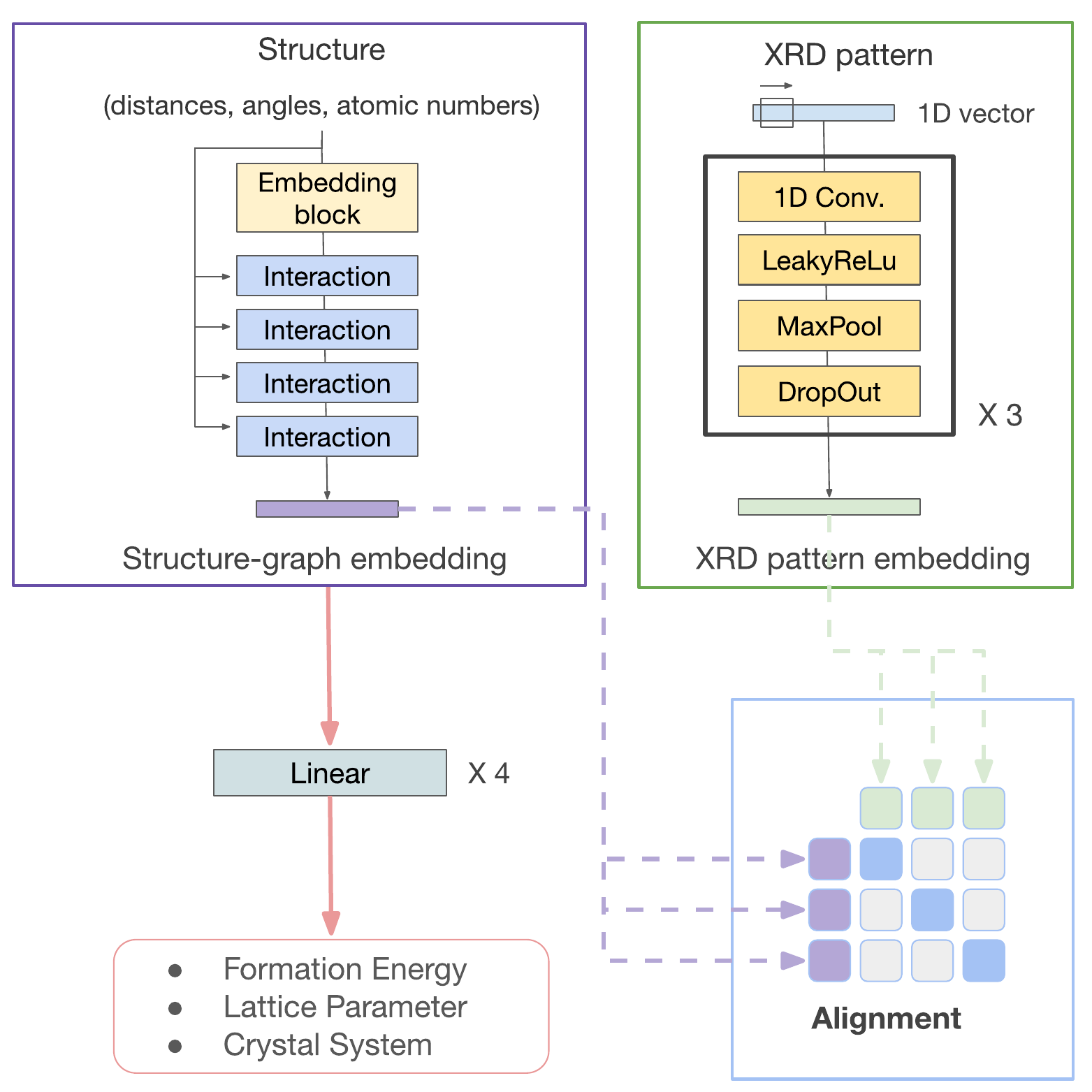}
  \caption{The best-performing model setup of the alignment experiment is shown. The structure graph encoded by GNN, and the XRD pattern encoded by CNN are trained with CLIP loss. The structure embedding was then utilized to perform downstream tasks.}
  \label{fig:align}
\end{figure}

In our first experiment, we investigated the effect of alignment on prediction accuracy of lattice parameters and crystal systems from structure graphs. Figure \ref{fig:align} illustrates the framework used for training and inference. This aims to enhance the single-modality structure graph by incorporating additional modalities. In order to estimate this effect, we calculate the mean absolute error (MAE) of all of the predicted lattice parameters, including lengths (Å) and angles (deg), and classification accuracy of crystal system (i.e. cubic, tetragonal, trigonal, etc.) of each crystal structure in the test set. The regression/classification tasks and inference steps are performed from an aligned embedding. We compared prediction accuracies across three scenarios: using only the XRD pattern as the input modality, using only the structure graph as the input modality, and using the structure graph as the input modality with XRD alignment applied. The results of this experiment are shown in Table \ref{tab:results}.

XRD patterns, when represented as a 1D vector as an input to a CNN-based encoder, achieve an MAE of 0.17 Å for lattice lengths and 4.5° for lattice angles in the test dataset. In comparison, structure graphs encoded by a GNN show higher prediction errors, with an MAE of 0.20 Å for lattice lengths and 6.1° for lattice angles. However, when structure graphs are aligned with XRD patterns, the MAE for lattice lengths decreases to 0.14 Å. Furthermore, lattice angle predictions from structure graphs with XRD alignment become comparable to those from XRD patterns alone, with both achieving an MAE of 4.5°.

For the task of crystal system classification, the accuracy of predictions from the XRD pattern, structure graph, and structure graph with XRD alignment are all close at 83.0\%, 83.2\%, and 83.1\%. However, there are distinct variations in prediction accuracy for crystal system classification and lattice parameter prediction across different crystal systems. There is a noticeable trend that the accuracy is lower for crystal systems with lower symmetry. The breakdown of accuracy for each crystal system can be found in SI Table 5.

\setlength{\extrarowheight}{3pt} 
\begin{table}[!htbp]
\centering
\caption{Alignment Experimental Results: Regression MAE and Classification Accuracy}
\label{tab:results}
\begin{tabular}{>{\raggedright\arraybackslash}m{3cm} >{\raggedright\arraybackslash}m{3cm} >{\centering\arraybackslash}m{2cm} >{\centering\arraybackslash}m{2cm} >{\centering\arraybackslash}m{2cm}}
\toprule
\textbf{Modality} & \textbf{Representation / Encoder} & \textbf{Lattice lengths (Å)} & \textbf{Lattice angles (deg)} & \textbf{Crystal Systems (\%)} \\
\midrule
Structure   & Graph / GNN (DimeNet++, Ref \citenum{gasteiger_fast_2022}) & 0.203 & 6.052 & 83.2\% \\
\hline
XRD pattern & 1D-vector / CNN & 0.173 & 4.487 & 83.0\% \\
Structure (pre-train w/ align) & Graph / GNN & 0.136 & 4.491 & 83.1\% \\
\bottomrule
\end{tabular}
\end{table}


\subsection{Multi-modal XRD and composition predictions improve accuracy versus structure graphs}

\begin{figure}[!hptb] 
  \centering
  \includegraphics[width=0.8\linewidth]{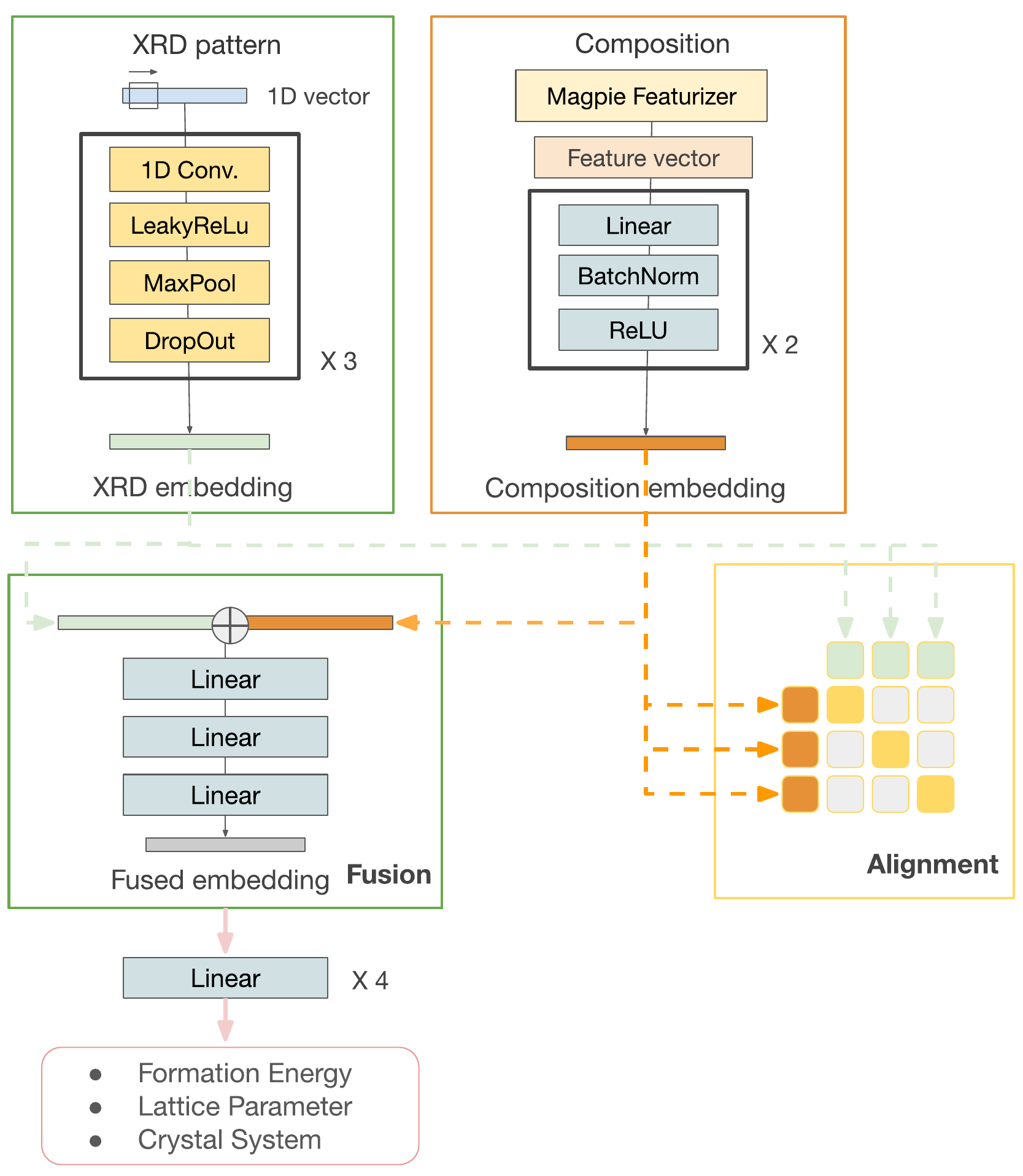}
  \caption{ The best-performing model setup from the fusion experiment is shown. The XRD pattern, encoded by a CNN, and the Magpie featurized composition, encoded by an MLP, are aligned by training with the CLIP loss. The fused embedding between the two is then utilized to perform downstream tasks.}
  \label{fig:fusion}
\end{figure}

In our second experiment, we demonstrate the results of the fusion and alignment of XRD and composition modalities, comparing to both the constituent single modalities and the alternative simulation based single modality of structure graphs. Alignment and fusion are applied together for regression and classification task, as shown in Figure \ref{fig:fusion}. This example is motivated by the desire to compare data streams that originate from simulation (structure graph) and experiment (composition, XRD). 

The results of MAEs of models corresponding to each of these scenarios are shown in Table 2. Structure graphs and XRD pattern as input modalities have been discussed previously, but we also note from Table 2 that formation energy as a regression target is significantly more accurately predicted in the case of structure graphs, with MAE 0.02 eV/atom compared to XRD patterns’ 0.31 eV/atom.

Composition alone is a very poor predictor of structural information with 0.33 Å and 9.6° MAE for lattice lengths and angles regression tasks. Similarly, composition alone struggles to accurately classify crystal systems, with a 58.5\% accuracy, significantly lower than the ~83\% of the previous examples.

With the single modalities as benchmarks, we now assess the effect of using a multi-modal approach that combines XRD and composition. Using the alignment and fusion of XRD and composition vectors, along with a CNN encoder, we observe 4.4° and 85\% in both the lattice angles MAE and crystal system classification accuracy, slightly improving on the 4.5° and 83\% of XRD single modality prediction alone. These values indicate significant improvement relative to the performance of composition alone, but are only marginal improvements on that of the XRD modality alone.

However, multi-modal XRD+composition alignment and fusion improves lattice parameter prediction accuracy significantly relative to that of any of our previously tested single-modality predictions. Specifically, an MAE of 0.13 Å is achieved on the regression task of predicting the average lattice parameter length, lower than the 0.17 Å or 0.33 Å of XRD and composition alone, and similarly lower than 0.20 Å of the structure graph.

\setlength{\extrarowheight}{3pt} 
\begin{table}[!hptb] 
\centering
\caption{Alignment + Fusion Experimental Results: Regression MAE and Classification Accuracy}
\label{tab:results2}
\begin{tabularx}{\textwidth}{>{\centering\arraybackslash}m{2.2cm} >{\centering\arraybackslash}m{1.9cm} >{\centering\arraybackslash}m{2.9cm} >{\centering\arraybackslash}m{1.7cm} >{\centering\arraybackslash}m{1.7cm} >{\centering\arraybackslash}m{1.7cm} >{\centering\arraybackslash}m{1.7cm} }
\toprule
\textbf{Accessibility} & \textbf{Modality} & \textbf{Representation / Encoder} & \textbf{Lattice lengths (Å)} & \textbf{Lattice angles (deg)} & \textbf{Formation Energy (eV/n)} & \textbf{Crystal system (\%)} \\
\midrule
Hard & Structure & Graph / GNN (DimeNet++, Ref \citenum{gasteiger_fast_2022}) & 0.203 & 6.052 & 0.023 & 83.2\% \\
\midrule
\multirow{3}{=}{\center{Easy}} & XRD & 1D-vector / CNN & 0.173 & 4.487 & 0.314 & 83.0\% \\
 & Composition & Elemental features / MLP & 0.330 & 9.639 & 0.059 & 58.5\% \\
 & XRD \& Composition (align+fuse) & 1D-vector / CNN, Elem. feat. / MLP & 0.128 & 4.385 & 0.088 & 85.1\% \\
\bottomrule
\end{tabularx}
\end{table}


\subsection{Multi-modal unsupervised embeddings clustering}

\begin{figure}[!hptb] 
    \centering
    \includegraphics[width=\textwidth]{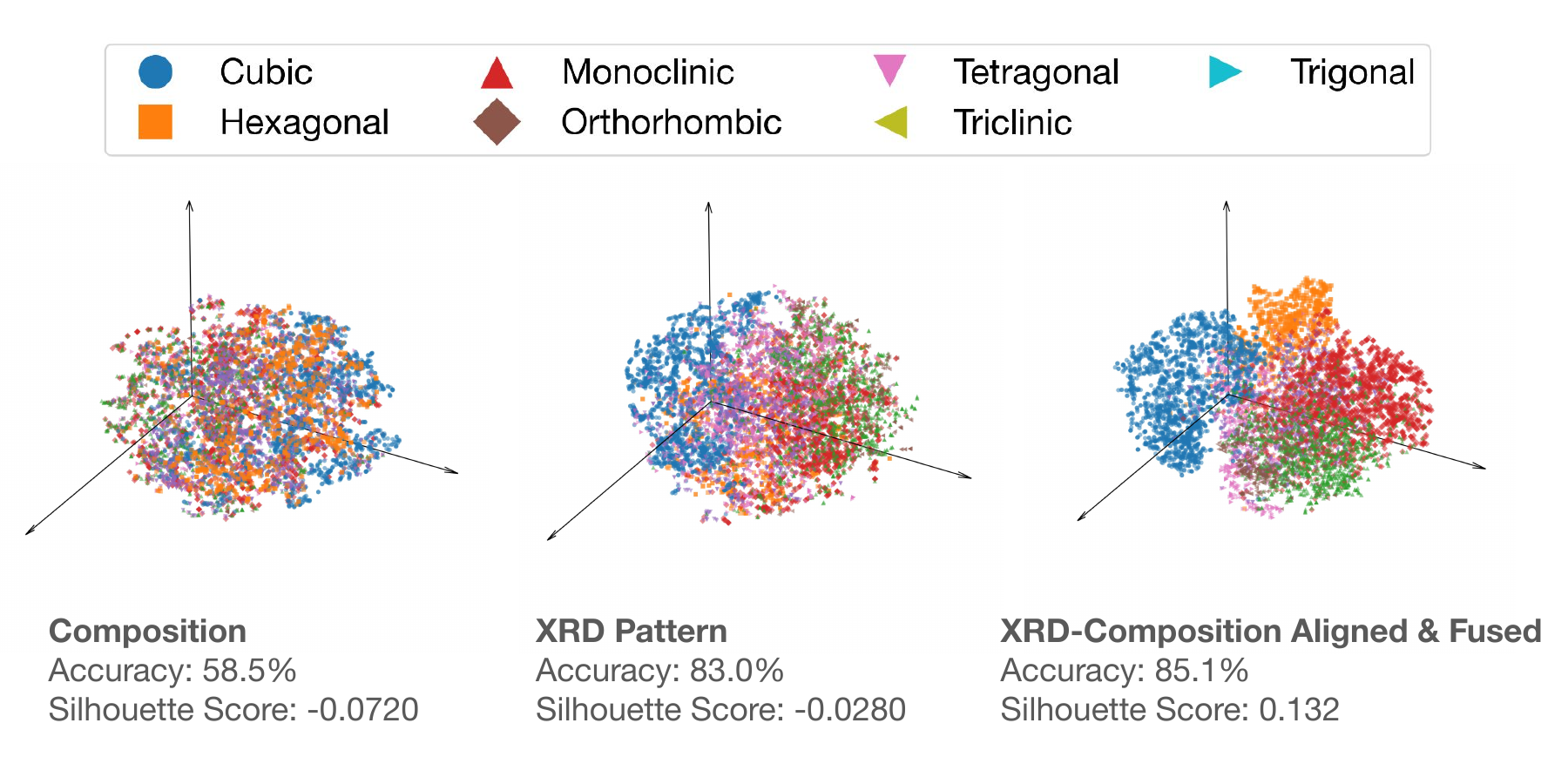}
    \caption{The clustering of latent space by crystal systems.}
    \label{fig:tsne}
\end{figure}

To further illuminate the effect of the multi-modality approach relative to a single modality, we also assessed 3D projections of our latent space grouped by crystal system. Figure \ref{fig:tsne} visualizes t-SNE-reduced 3D representations of the latent space learned from XRD patterns alone, composition alone, and the aligned and fused combination of both. Each data point is colored by crystal system. Visual inspection of the multi-modal case suggests that the latent space of learned embeddings has improved clustering w.r.t crystal system. We further quantified clustering of embedding space w.r.t crystal systems using silhouette scores, which increased from -0.028 (XRD), -0.072 (composition) to 0.132 (alignment and fusion).

We further analyze the classification accuracy for each crystal system present in the dataset to understand the model performances for different symmetries. Due to the significant under-representation of triclinic systems, with a train ratio of 4.2\% and a test ratio of 4.1\%, we exclude them from the discussion. As indicated in Table \ref{table:crystal_system_ratio}, the crystal systems are listed in descending order of symmetry. We observe a strong correlation between symmetry and classification accuracy. The cubic system, which has the highest symmetry, achieves the highest accuracy at 98.1\%. Meanwhile, systems with lower symmetry, such as trigonal and monoclinic, exhibit lower accuracies of 80.0\% and 78.1\%, respectively.

\begin{table}[!htbp]
  \centering
  \caption{Crystal System Ratios in Train and Test Datasets. The accuracies are from the XRD-Composition Aligned and Fused Model.}
  \label{table:crystal_system_ratio}
  \resizebox{\textwidth}{!}{%
  \begin{tabular}{lccc}
    \toprule
    \textbf{Crystal System} & \textbf{Train Ratio} [\%] & \textbf{Test Ratio} [\%] & \textbf{Classification Accuracy} [\%] (\(\uparrow\)) \\
    \midrule
    Cubic         & 22.9  & 23.8 & 98.1 \\
    Hexagonal     & 19.4  & 19.1 & 83.6 \\
    Trigonal      & 11.3  & 10.7 & 80.0 \\
    Tetragonal    & 16.7  & 17.3 & 83.3 \\
    Orthorhombic  & 16.7  & 19.1 & 82.0 \\
    Monoclinic    & 15.0  & 14.2 & 78.1 \\
    Triclinic     & 4.2   & 4.1  & 72.6 \\
    \bottomrule
  \end{tabular}%
  }
\end{table}

\section{Discussion}

We begin our discussion with a commentary on the modalities we examine in this work, explaining the heuristics of how the modalities are understood in domain terms. In this, we intend to provide machine-learning practitioners in particular a more thorough background in the scientific practice associated with each of the tasks from our work.

Structure graphs are perhaps the richest in chemical information, where the nodes and edges of the structure graph capture information about atomic identity and the connectivity between atoms. This implies that the structure modality is likely to capture the underlying electronic information from which the atom identity and connectivity are derived. Composition, on the other hand, retains coarse-grained information about atomic identity as a weighted average, keeping the heuristic connection to periodic table-level information, often also drawing from properties of pure elemental compounds that may not be contained in the periodic table. Composition, of course, does not include the local information encoded in the edges of the structure graph. Thus, it seems intuitive that the structure graph and composition typically should perform well on tasks more closely related to electronic structure like formation energy and band gap.


XRD patterns, on the other hand, because they are related to the constructive interference of X-rays scattered from periodically arranged atoms, are closely related to the symmetry of crystalline materials. Thus, they are the most routinely used diagnostic tool for labeling crystal structures. Similarly, it is intuitive that the XRD pattern modality should perform well on lattice parameter prediction tasks and crystal system classification. This heuristic understanding of the relevance of chemical and structural information for structure graphs, composition, and XRD patterns is reflected in the performance of these modalities in the prediction tasks presented herein.

We also note that the modalities have different levels of accessibility to practicing materials scientists, who are often further distinguished by expertise in simulation or experiment. Simulation practitioners regularly work with structure graphs, particularly with the advent of GNN-based ML models for forward prediction of materials properties from crystal structure. However, structure graphs are not typically used directly in provisioning simulations, but are obtained from more explicit representations of crystal structure that include atomic coordinates, atomic identity, and lattice length and angle information. They are primarily used as an intermediate representation that is more amenable to neural-network based prediction. However, full crystal structures from which structure graphs are usually derived, can be very difficult to solve in the first place, and are typically derived effectively as labels of experimental information like XRD patterns, microscopic images, or other spectra. XRD patterns and composition, on the other hand, are far more commonly used and accessible to experimentalists. A typical interface of simulation and experimental involves accessing simulation derived properties from structures obtained by solving XRD patterns, wherein solving structure from XRD patterns is an under-constrained problem and is an active area of research. Our work shows that integrating multiple modalities is promising to potentially capture context/constraints describing a material that could result in enhanced prediction of structural features (lattice parameters, symmetry, etc.) or properties from experimental modalities.

In future work, we see potential in exploring more sophisticated fusion strategies, which have shown success in multiple configurations, including the stage of fusion (early fusion vs. late fusion), fusion block design, and training strategies (pre-training vs. training from scratch). However, optimizing encoders for XRD pattern-like data remains a challenge, as high-quality, easily accessible encoders are limited. While multiple advanced pre-trained Graph Neural Networks (GNNs) and composition based models, the field lacks a universal pre-trained model for XRD pattern-like data. Addressing this gap could significantly enhance the performance of future multi-modal model applications.

Moreover, adapting our current methods to work with experimentally measured X-ray diffraction (XRD) data, rather than the simulated data used in this study, poses an additional challenge. This adaptation is crucial for real-world applicability. Data augmentation techniques have shown promise in enhancing model performance during phase transitions \cite{sun2023application} and are on our agenda for future exploration. By incorporating these advanced strategies and addressing the existing gaps, we aim to improve the robustness and applicability of our models in practical settings.

\section{Conclusion}
In this work, we present UniMat, a methodology to Unify Materials Embeddings through Multi-modal Learning, and provide evidence that this methodology is capable of a) enhancing a single modality through integration with additional modalities. Specifically, we showed that structure graph modality aligned with XRD significantly improves lattice length prediction, b) we also demonstrated that modalities that are less sensitive to a specific property may not be able to make a significant impact through an alignment task, however, fusion of these "weaker" modalities with other modalities can significantly enhance predictive performance. Specifically, we demonstrated that aligning and fusing XRD and composition significantly improves the lattice length prediction while nearly retaining the performance of the best modality for other properties, c) we also demonstrated that experimentally accessible modalities can rival the performance of simulation-based modalities, thus providing an effective platform for prediction purely from practically relevant experimental modalities. Specifically, we demonstrate that experimentally accessible modalities such as XRD and composition can be used to capture lattice lengths, angles, and crystal systems as effectively or better than structure graphs. Finally, we highlight that these successful demonstrations present a pathway for enabling real-world accelerated materials discovery using multi-modal AI applications in materials science.

%




\bibliography{references}
\end{document}